# GRADIENT VECTOR FLOW MODELS FOR BOUNDARY EXTRACTION IN 2D IMAGES

Gilson A. Giraldi, Leandro S. Marturelli, Paulo S. Rodrigues
LNCC–National Laboratory for Scientific Computing
Av. Getulio Vargas, 333, 25651-070 Petrópolis, RJ
Brazil
{gilson,schaefer,pssr}@lncc.br

**ABSTRACT**

The Gradient Vector Flow (**GVF**) is a vector diffusion approach based on Partial Differential Equations (**PDEs**). This method has been applied together with snake models for boundary extraction medical images segmentation. The key idea is to use a diffusion-reaction PDE to generate a new external force field that makes snake models less sensitivity to initialization as well as improves the snake's ability to move into boundary concavities. In this paper, we firstly review basic results about convergence and numerical analysis of usual GVF schemes. We point out that GVF presents numerical problems due to discontinuities image intensity. This point is considered from a practical viewpoint from which the GVF parameters must follow a relationship in order to improve numerical convergence. Besides, we present an analytical analysis of the GVF dependency from the parameters values. Also, we observe that the method can be used for multiply connected domains by just imposing the suitable boundary condition. In the experimental results we verify these theoretical points and demonstrate the utility of GVF on a segmentation approach that we have developed based on snakes.

## 1. Introduction

In image processing, the use of diffusion schemes that can be formulated through Partial Differential Equations (**PDEs**) is an useful practice. A remarkable work in this area came from the observation that the Gaussian filtering can be seen as the fundamental solution of the (linear) heat equation [8]. Then, Perona and Malik consider a non-linear heat equation and proposed their anisotropic (nonlinear) diffusion method [10]. Since then, PDE approaches have been used in multiscale techniques [6], image restoration, noise reduction and feature extraction [1,3]. From the viewpoint of snake models, these methods can be used to improve the convergence to the desired boundary [12]. Multivalued image diffusion schemes were also explored and its advantages have been reported in the literature [1]. The Gradient Vector Flow (GVF) model [18], Chromaticity Diffusion [13] and total variation methods [4] are known examples among other ones [1]. Also, Anisotropic diffusion has been extended for multivalued signals [11]. These methods are based on PDEs, which may be derived from variational problems. As usual in image processing applications, an initial value problem is associated to the PDE [3]. Thus, starting from an initial field it is constructed a family $\{v(x,t)\ ;\ t>0\}$ representing successive versions of the initial one. As $t$ increases we expect that $v(x, t)$ changes into a more and more suitable field and converges to the steady-state solution. More fundamental for this work, the convergence to the steady-state solution depends on the convexity properties of the variational problem [19]. In this paper, we focus on the GVF methods. We firstly review basic results about the global optimality and numerical analysis of GVF [17, 18, 19]. These results rely on the assumption that the solution of the variational problem belongs to a Sobolev space (functions with *generalized* derivatives [5]). However, in the problems of interest the discontinuities in the images are significant and important features. Unfortunately, classical Sobolev spaces do not allow one to account for discontinuities since the gradient of a discontinuous function $f$ has to be understood as a measure [3]. This theory can be precisely formulated in the context of functions of *bounded variation*, which is out of the proposal of this paper. Instead, in this work we take a more practical viewpoint based on a finite-difference version of the GVF. Also, analytical analyses of the time derivative of the GVF equation (velocity equation) offer elements to discuss the sensitivity of the model to parameter selection and properties of the steady-state solution. Finally, we observe that it is straightforward to extend the GVF to multiply connected domains. These are the contributions of this paper. Their practical consequences are shown in the tests that we made.

## 2. Gradient Vector Flow (GVF)

GVF is a vector diffusion method that was introduced in [18] and can be defined through the following diffusion-reaction equation [19]:

$$\frac{\partial v}{\partial t} = \nabla.(g\nabla v) + h(\nabla f - v), \quad v(x,0) = \nabla f \qquad (1)$$

where $f$ is a image gradient function (for example, $P = -\|\nabla I\|^2$), and $g(x)$ and $h(x)$ are nonnegative functions,

defined on the image domain. The field obtained by solving the above equation is a smooth version of the original one, which tends to be extended very far away from the object boundaries. When used as an external force for deformable models, it makes the methods less sensitive to initialization [18] and improves their convergence to the object boundaries, as we shall see later. Expression (1) can be derived from a variational problem. To see this, let us denote a $x \in \mathbb{R}^2$ by $x = (x^1, x^2, ..., x^n)$, a scalar function at $x$ by $f(x) = f(x^1, x^2, ..., x^n)$ and a vector function at $x$ by $v(x) = (v^1(x), x^2(x), ..., x^n(x))^T$. The gradient of $f$ gives a vector field and the gradient of v yields to a tensor field, given by:

$$\nabla f = \left(\frac{\partial f}{\partial x^1}, \frac{\partial f}{\partial x^2}, ..., \frac{\partial f}{\partial x^n}\right)^T \; ; \; \nabla v = \left(\frac{\partial v_i}{\partial x^j}\right)_{i,j=1,...,n} \quad (2)$$

respectively. As usual, the Euclidean inner product between vector functions u and v is denoted by $u \cdot v$ and the inner product between tensors $T$ and $S$ as $T.S = \sum_{i,j=1}^{n} T_{ij} S_{ij}$. It is also supposed that all these functions are defined in a bounded domain $\Omega \subset \mathbb{R}^n$ with $\partial \Omega$ as its boundary. The n-dimensional GVF is defined as the vector function $v(x)$ in a subset of the Sobolev space denoted by $W_2^2(\Omega)$ which minimizes the following functional [19]:

$$J(v) = \int_{\Omega} \left(g \|\nabla v\|^2 + h \|\nabla f - v\|^2\right) dx \quad (3)$$

where $g$ and $h$ are nonnegative functions defined on, $\Omega$, $\|\nabla f - v\|$ and $\|\nabla v\|$ are the norms for vectors and tensors given by $\sqrt{\nabla v . \nabla v}$ and $\sqrt{(\nabla f - v).(\nabla f - v)}$, respectively. The argument of the integral (3), denoted as usual by $L(x, v, \nabla v)$, is called the *Lagrangian* of the variational problem. Following, we review the work [19], where it is established the necessary and sufficient conditions for globally minimizing the variational problem. The sufficient conditions, the known Euler-Lagrange Equations, give the steady-state solution of problem (1). Therefore, let us present some definitions and a fundamental proposition.

**Definition 1:** For $y, v \in W_2^2(\Omega)$, the Gateaux variation of $J$ at $y$ in the direction $v$ is defined by:

$$\delta J(y, v) = \lim_{\varepsilon \to 0} \frac{J(y + \varepsilon v) - J(y)}{\varepsilon}$$

**Definition 2:** A real-valued functional $J$ defined on $W_2^2(\Omega)$ is said to be convex on $W_2^2(\Omega)$ provided that when $y$ and $y + v \in W_2^2(\Omega)$ then $\delta J(y, v)$ is defined and $J(y + v) - J(y) \geq \delta J(y, v)$, with equality if and only if $v = 0$.

**Fundamental Proposition:** If $J$ is convex on $W_2^2(\Omega)$, then each $y_0 \in W_2^2(\Omega)$ for which $\delta J(y_0, v) = 0, \forall v \in W_2^2(\Omega)$ minimizes $J$ on $W_2^2(\Omega)$. Thus, it follows that we must to show that the GVF functional is convex, according to Definition 2 and then to set sufficient conditions to assure that there is a $y_0$ that satisfies the Fundamental Proposition just stated.

**Lema 1:** The GVF functional is strictly convex when $g$ and $h$ are not both zero at any $x \in \Omega$.
Proof: Following Definition 2, we should proof that $J(y + v) - J(y) \geq \delta J(y, v) \; \forall u, v \in W_2^2(\Omega)$. Thus, by substituting the definition of GVF, a simple algebra shows that (see [19] for details):

$$J(y + v) - J(y) = \int_{\Omega} [g \|\nabla u\|^2 + 2g \nabla v . \nabla u + h \|u\|^2 + \\ + 2h(\nabla f - v).u] dx \quad (4)$$

By substituting $u \to \varepsilon u$ in this expression and applying Definition 1 for the Gateaux variation of GVF, it is straightforward to show that:

$$\delta J(y, v) = \int_{\Omega} [2g \nabla v . \nabla u + 2h(\nabla f - v).u] dx \quad (5)$$

Hence, from Expressions (4) and (5) we have:

$$J(y + v) - J(y) - \delta J(v, u) = \int_{\Omega} [g \|\nabla u\|^2 + h \|u\|^2] dx$$

Besides, if functions $g$ and $h$ are not both zero at any $x \in \Omega$ the equality holds if and only if $u = 0$, which complete the proof. The sufficient conditions are stated bellow:

**Lema 2:** In the conditions of the Lema 1, each $v \in W_2^2(\Omega)$ satisfying:
$$\nabla .(g \nabla v) = h(\nabla f - v), \quad (\nabla v).N = 0, \quad \text{on } \Omega,$$
where $N = (N^1, ..., N^n)^T$ is the normal of the boundary $\partial \Omega$ and $(\nabla v).N = \left(\sum_{i=1}^{n} \frac{\partial v^1}{\partial x^i} N^i, ..., \sum_{i=1}^{n} \frac{\partial v^n}{\partial x^i} N^i\right)^T$ minimizes the GVF functional on $W_2^2(\Omega)$. The first expression above is just the Euler-Lagrange equations for the GVF.
Proof: To prove this Lema, we start from the Expression (5) and rewrite the first term of that integral as:

$$\int_{\Omega} \nabla .[2g(\nabla v)u] dx - \int_{\Omega} [\nabla .(2g \nabla v)] .u dx = \\ \int_{\partial \Omega} [2g(\nabla v)u] N dS - \int_{\Omega} [\nabla .(2g \nabla v)] .u dx,$$

where $dS$ means the element of integration on $d\Omega$. However, this expression can be simplified by observing that the first term vanishes due to the boundary condition:

$$\int_{\partial\Omega}[2g(\nabla v)u]NdS = 2\int_{\partial\Omega}[g(\nabla v)^T N].udS = 2\int_{\partial\Omega}[g(\nabla v)N]udS = 0,$$

where we have used the fact that $\nabla v$ is symmetric. Thus, we finally have:

$$\delta J(v,u) = \int_{\Omega}[2h(\nabla f - v) - \nabla.(2g\nabla v)].udx.$$

Hence, if $\delta J(v,u) = 0$, $\forall u$, we obtain the Euler-Lagrange equations for the GVF. If $v$ is also considered as a function of time, then the solution for the above equations becomes the steady-state one for the initial value problem (1). Now, we summarize the results presented in [18, 17] about numerical solutions of the GVF model in Expression (1), when $g$ is a constant. For simplicity, we restrict to the bidimensional case and simplify notations by changing $(x^1, x^2)$ by $(x,y)$. Such as [18], we will consider the following explicit, finite-difference, scheme:

$$\frac{\partial v^k}{\partial y} \approx \frac{1}{\Delta t}\left((v^k)_{i,j}^{n+1} - (v^k)_{i,j}^{n}\right), \quad (6)$$

$$\nabla^2(v^k) \approx \frac{1}{\Delta x \Delta y}\left((v^k)_{i+1,j} + (v^k)_{1,j+1}\right) + \frac{1}{\Delta x \Delta y}\left((v^k)_{i-1,j} + (v^k)_{1,j-1} - 4(v^k)_{i,j}\right) \quad (7)$$

where k = 1, 2. By substituting these expressions in Equation (1), with $g$ constant, we get the following system:

$$(v^k)_{i,j}^{n+1} = (1 - h_{i,j}\Delta t)(v^k)_{i,j}^{n} + c_{i,j}^k \Delta t + r\left((v^k)_{i+1,j}^n + (v^k)_{i,j+1}^n + (v^k)_{i-1,j}^n\right) + r\left((v^k)_{i,j-1}^n - 4(v^k)_{i,j}^n\right), \quad (8)$$

where $c^1, c^2$ and $r$ are functions given by:

$$c^1(x,y) = h(x,y)\frac{\partial f}{\partial x}, \quad c^2(x,y) = h(x,y)\frac{\partial f}{\partial y}, \quad (9)$$

$$r = \frac{g\Delta t}{\Delta x \Delta y} \quad (10)$$

Provided that $h, c^1$ and $c^2$ are bounded, this scheme is stable if $r < 1/4$ [2]. Besides, Expressions (6) and (7) converge to $\partial u/\partial t$ and $\nabla^2 u$ in the limit $(\Delta t, \Delta x, \Delta y) \to (0,0,0)$. Thus, the numerical scheme is consistent and stable. Henceforth, according to the fundamental *Equivalence Theorem of Lax* [7], a sufficient condition for the numerical solution to converge to the exact one is the initial value problem (1) to be well posed. However, variational methods in computer vision are, in general, ill-posed (too sensitive to the initial and boundary conditions [7]), if regularization constraints or multiscale techniques [3] are not used. Besides, the effects of the functions $h$ and $g$ have not been deeply discussed yet in the GVF literature. They control the number of time steps needed to achieve the usual termination condition given by:

$$\max_{i,j}\left\|v_{i,j}^{k+1} - v_{i,j}^{k}\right\| < \delta \quad (11)$$

where $\delta$ is a pre-defined parameter and $\|\cdot\|$ can be the usual 2-norm. These questions and their importance for boundary extraction are the starting point for our work as follows.

## 3. Numerical Analysis

In the problems of interest the discontinuities in the images are significant and important features. As a consequence, it was experimentally observed that if the GVF, given by Expression (1), is applied, the numerical solution sometimes diverges from the steady-state solution. To bypass such problem, we can apply multiscale methods [14] or even impose the constraint $\|\nabla f\| \leq T$, where T can be set by intensity histogram analysis, in order to keep bounded variation functions (see http://iacl.ece.jhu.edu/projects/gvf/faq.html). However, we observe in the experiments that even such procedure may fails. How to address this problem? We are going to consider this question from a practical viewpoint letting a more rigorous analysis, based on the theory of functions of *bounded variation* [3] for further works. The explicit finite-difference scheme of Section 2 can be formally written as:

$$v(n+1) = v(n) + \Delta t\left(g\Delta v(n) + h(\nabla f - v(n))\right), \quad (12)$$

$$v(0) = \nabla f; \quad (13)$$

where $v(n)$ means the (discrete) vector field obtained after n-iterations. In this compact notation, the results for the first 4 iterations can be written as follow:

$$v(1) = \nabla f + \Delta t g \Delta(\nabla f),$$

$$v(2) = \nabla f + \Delta t g (2 - h\Delta t) \Delta(\nabla f) + (\Delta t g)^2 \Delta^2(\nabla f),$$

$$v(3) = \nabla f + \Delta t g \left(3 - 3h\Delta t + (h\Delta t)^2\right) \Delta(\nabla f) +$$
$$(\Delta t g)^2 (3 - 2h\Delta t) \Delta^2(\nabla f) + (\Delta t g)^3 \Delta^3(\nabla f),$$

$$v(4) = \nabla f + \Delta t g \left(4 - 6h\Delta t + 4(h\Delta t)^2 - (h\Delta t)^3\right) \Delta(\nabla f) +$$
$$(\Delta t g)^2 \left(6 - 8h\Delta t + 3(h\Delta t)^2\right) \Delta^2(\nabla f) +$$
$$(\Delta t g)^3 (4 - 3h\Delta t) \Delta^3(\nabla f) + (\Delta t g)^4 \Delta^4(\nabla f).$$

where $\Delta^2(\nabla f) = \Delta(\Delta(\nabla f))$, $\Delta^3(\nabla f) = \Delta(\Delta^2(\nabla f))$, and so on. These expressions suggest the two necessary conditions for the numerical solution $v(n)$ to converge to the numerical solution of the Euler-Lagrange equation of GVF are: $g\Delta t < 1$; $h\Delta t < 1$.

The first one can be obtained as a consequence of the condition $r < 1/4$, where $r$ is given by Expression (10). For instance, if $\Delta x = \Delta y = 1$, as usual for digital images, we get $g\Delta t < 1/4$, and so, the first constraint is satisfied. From the finite-difference and finite elements literature [2], it is known that the method presents numerical difficulties for small values of $(g/h)$. Besides, as a consequence of it ($h < g$), the constraint $h\Delta t < 1$ is satisfied too. Thus, the numerical scheme must agree with the following constraints:

$$\frac{g\Delta t}{\Delta x \Delta y} < \frac{1}{4}, \quad g\Delta t < 1, \quad h\Delta t < 1, \quad h < g \quad (14)$$

Besides, parameters $g$ and $h$ control the number of time steps need to achieve the usual termination condition given by Equation (11). This is demonstrated next.

## 4. Revised GVF

In this section, we propose and demonstrate new GVF properties that will be explored in the experimental results (Section 6). For simplicity, we suppose $h$ and $g$ as constants in the GVF model, given by the Equation (1). Now, we consider the effect of the parameter $h$ in the convergence of the solution of (1) to the steady-state one, and consequently, the number of time steps need to achieve condition of Equation (11).

**Property 1:** If we increase $h$ we decrease the rate of convergence to the steady-state solution. The same for $g$.
Proof: If we take the time derivative of the GVF Equation (1), we get an equation for the vector field velocity, given by:

$$\frac{\partial}{\partial t}(v_t) = g\Delta(v_t) - h(v_t), \quad (15)$$

where $v_t = \partial v/\partial t$. $v_t = \partial v/\partial t$. If we multiply both sides of expression (15) for $v_t$ and integrate by parts we will obtain the following expression:

$$\frac{1}{2}\frac{\partial}{\partial t}\int_\Omega \|v_t\|^2 \, d\Omega = -g\int_\Omega \|\nabla v_t\|^2 \, d\Omega - h\int_\Omega \|v_t\|^2 \, d\Omega, \quad (16)$$

which implies that the field *energy* also decreases faster when $g$ (or $h$) increases. Thus, we are increasing the rate of convergence. Such conclusion will be verified in the experimental results. Finally, we observe that we can apply GVF to *multiply connected* domains.

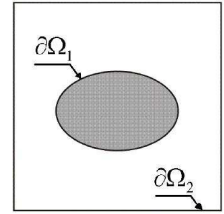

Figure 1: Multiply connected domain for GVF.

In order to verify this point, let us suppose that the boundary $\partial\Omega$ in Expression (3) has two disconnected components, that is, $\partial\Omega = \partial\Omega_1 \cup \partial\Omega_2$ and $\partial\Omega_1 \cap \partial\Omega_2 = 0 \Rightarrow \Omega = \Omega_1 - \Omega_2$ (see Fig. 1). Then, we can define the GVF as:

$$J(v) = \int_{\Omega = \Omega_1 - \Omega_2} L dx = \int_{\Omega_1} L dx - \int_{\Omega_2} L dx \quad (17)$$

The Lagrangian $L$ is that one presented in the functional (3). This definition assumes that the integral is defined on 2, which is not restrictive for our applications. By doing this, we can straightforward generalize the results presented in Section 2.

## 5. Frequency Domain Analysis

In this section we offer a Fourier analysis of GVF to show the low-pass nature of GVF. Taking the Fourier transform of the steady-state equation $(\partial v/\partial t = 0)$, in Expression (1), supposing $h$, $g$ constants, we find the following expressions:

$$-g\left[\omega_1^2 V_i + \omega_2^2 V_i\right] + h\left[F_i - V_i\right] = 0; \quad i = 1, 2 \quad (18)$$

where $V_1 = V_1(\omega_1, \omega_2)$ and $V_2 = V_2(\omega_1, \omega_2)$ are the Fourier transform of $v_1$ and $v_2$ respectively, and $(F_1, F_2)$ is the Fourier transform of $\nabla f(x, y) = (f_x(x, y), f_y(x, y))$ (we have used the

properties of the Fourier transform and derivatives [5]). These expressions can be rewritten as:

$$V_i(\omega_1, \omega_2) = \frac{F_i(\omega_1, \omega_2)}{\frac{g}{h}(\omega_1^2, \omega_2^2) + 1}, \quad i = 1, 2 \quad (19)$$

The analysis of these expressions gives the effect of the GVF in the frequency domain, for $h$ and $g$ constants. The practical consequences will be discussed next. We see that the steady-state solution is a low-pass filtering of the initial field, with the filter:

$$H(\omega_1, \omega_2) = \frac{1}{\sigma(\omega_1^2 + \omega_2^2) + 1}, \quad \sigma = \frac{g}{h} \quad (20)$$

Henceforth, in the space domain, the solution is smoothed version of the original field. Thus, we get the following property:

**Property 2:** The steady-state solution of GVF is such that $\left((v^1)^2 + (v^2)^2\right) \leq \|\nabla f\|^2$. The above development resembles the one that appears in scale space tracking based on deformable sheet models [15]. As a consequence, we can apply such approach to analyze GVF multiresolution/multiscale schemes [9,14] as well as to design new ones. These are further directions for our work.

## 6. Experimental Results

In this section we perform experimental results with GVF plus a simple snake model. We analyze GVF models respect to (a) sensitivity to the initial intensity field; (b) sensitivity to the functions/parameters $g$ and $h$; (c) domain changing. To demonstrate the GVF utility, we use a simple snake model as follows. The snake is a set of $N$ points $\{v_i = (x_i, y_i), i = 0, ..., N-1\}$ connected to form a closed contour. These points are called *snaxels*. The snake will be evolved based on a tensile (smoothing) force ($B_i$), given by:

$$B_i = \left(v_i - \frac{1}{2}(v_{i-1} + v_{i+1})\right) \quad (21)$$

and on the external (image) force ($f_i$) given by the GVF result, as proposed in [18]. Hence, we update the snaxels according to the following evolution equation:

$$v_i^{(t+\Delta t)} = v_i^t + h_i \left(b_i B_i^t + \gamma_i f_i^t\right) \quad (22)$$

where $b_i$ and $\gamma_i$ are scale factors and $h_i$ is an evolution step. A termination condition is defined based on small deformation, that is, $\max_i \left\| v_i^{(t+\Delta t)} - v_i^t \right\| < \varepsilon$. Fig. 2.(a)-(c) pictures the images that we will use in the following tests.

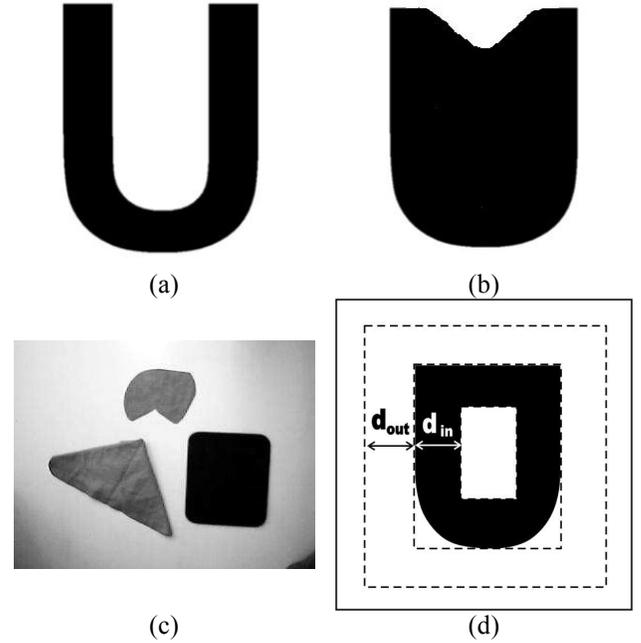

(a)          (b)

(c)          (d)

Figure 2: (a)-(c) Images to be used in this section, with resolutions, 400×400, 400×400 and 638×478, respectively. (d) Inner and outer window definition for Section 6.2.

In what follows, NI means the number of iterations of the numerical scheme and $T$ is the threshold used to constrain the initial intensity field ( $\max \|\nabla f\| \leq T$ ).

### 6.1 Intensity Field and Parameters

First of all, we consider a non-thresholded initial field ($T = \infty$). Table 1 reports the performed experiments for $h$, $g$ constants in Expression (1). The test image is pictured on Fig. 2.a.

| $g$ | 0.2 | 0.7 | 1.0 | 2.0 | 2.0 | 2.0 | 2.0 |
|-----|-----|-----|-----|-----|-----|-----|-----|
| $h$ | 0.01 | 0.01 | 0.01 | 0.01 | 0.02 | 0.05 | 0.1 |
| NI | 3406 | 2377 | 2072 | 1392 | 863 | 458 | 77 |

Table 1: Results for $\delta = 0.0001$ in Expr. 11 and $T = \infty$.

We can verify the sensitivity with respect to the parameter $g$. For instance, for $h = 0.01$, the number of iterations goes from 3406 (for $g = 0.2$) to 1392 when $g = 2.0$. The sensitivity of parameter $h$ (Property 1) is also noticed by observing the number of iterations. For example, for $g = 2.0$ we observe that the number of iterations goes from 1392 to 77. We must emphasize that the time step $\Delta t$ was kept unchanged for these experiments ($\Delta t$ =0.12). We choose the values for *g* based on the literature [15] and *h* based on the constraints in Expressions (14). An important point is the *quality* of the generated field. It must allow snake convergence insensitive to initialization. We can observe in Figure 3.a that the generated field is extended for almost the whole image. Another point with respect to the quality of the generated field is the GVF field topology inside the "U" cavity. The ideal situation is that in which the GVF result has no singular points (points in which $v = 0$) inside the cavity. Unfortunately, it is not the case in Fig. 3.a. This field is not able to push a snake into the boundary concavity, because the GVF field directions are almost orthogonal to the "U" walls inside the cavity. Thus, there is no component force towards the cavity.

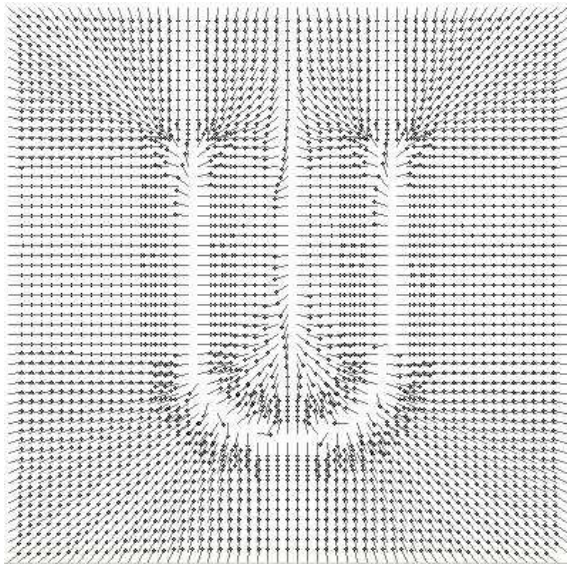

(a)

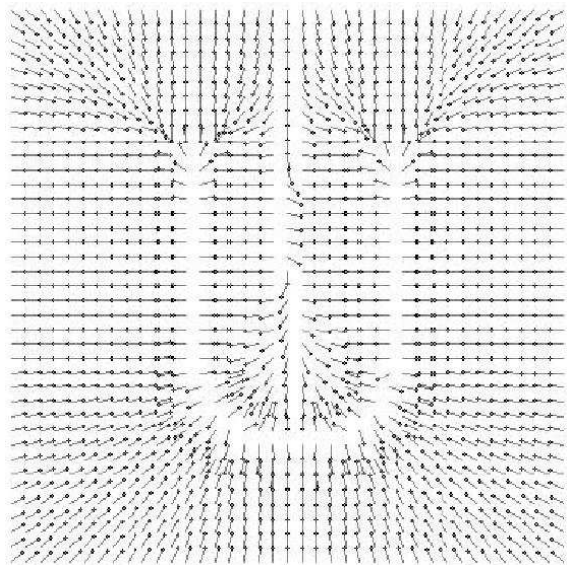

(b)

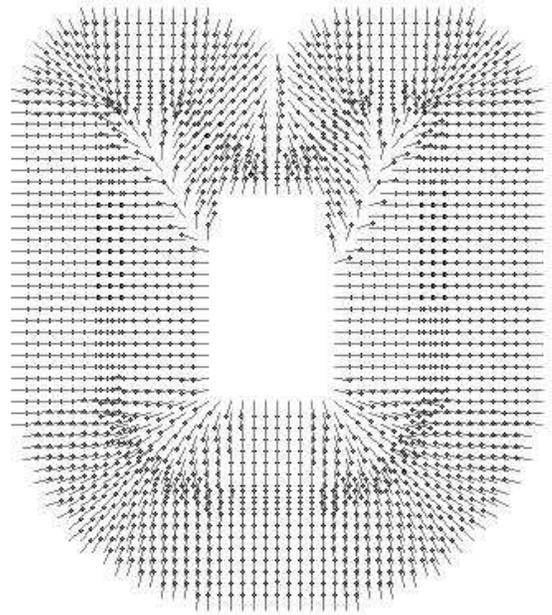

(c)

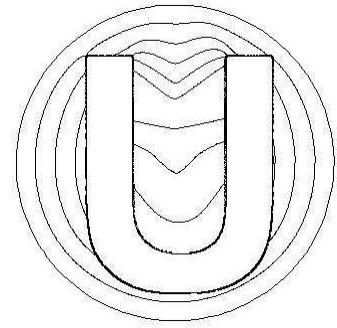

(d)

Figure 3: (a) GVF with $g$=2.0, $h$=0.02, $\delta$ =0.0001, $T = \infty$, for the image 2.a. (b) GGVF for $\delta$ =0.01, $K$ =100. (c) Multiply connected domain and GVF result for $g$=2.0, $h$=0.02, $T$=4 and $d_{in}$ = 60; (d) Snake evolution.

The Generalized GVF (**GGVF**), given by [17]:

$$\frac{\partial v}{\partial t} = g \Delta v + (1-g)(\nabla f - v); \quad (23)$$

$$g = \exp\left(-\frac{\|\nabla f\|^2}{K^2}\right) \quad (24)$$

with $K = 100$, performs better with respect to this requirement, as already observed in [17]. Fig. 3.b, shows the GGVF result and Fig. 3.d pictures the (desirable) snake evolution. Despite of this ability of GGVF, its computational cost can be very high. In this case, the number of iterations was 2390 with $\delta$ =0.05. Table 2.a reports some results when the initial intensity field is limited $\left(\|\nabla f\| \leq T < \infty\right)$. Once the generated field has

Property 2, it is enough limit just the initial intensity. Table 2 shows that the number of iterations got larger as *T* was increased.

| T | 1 | 4 | 7 | 10 | 40 | 80 |
|---|---|---|---|---|---|---|
| NI | 216 | 319 | 372 | 409 | 538 | 594 |

Table 2: Results for *g* = 2.0, *h* = 0.02 and $\delta = 0.0001$ in Expression 11.

## 6.2 Changing Domain Boundary

In this section, we consider the GFV behavior when the boundary of the domain is changed. Firstly, we consider the GVF sensitivity to domain reduction by taking an outer window according to Fig. 2.d. We observe from Table 3.a that the number of iterations remains insensitive until $d_{outer} = 50$. This effect was also observed for *T* = 1 as well as when using the image in Fig. 2.b. We should study this behavior in further works. Now, let us consider the method for a multiply connected domain. We observe that the convergence remains almost insensitive to the size of the inner window. Thus, once the computational cost of GVF is $O(n \cdot m)$, where $n \cdot m$ is the image size, and the number of iterations remained insensitive to domains size reduction, the computational cost was always reduced also, which is an important feature.

| $\delta$ | $d_{out}$ | Max | 60 | 50 |
|---|---|---|---|---|
| 0.0005 | NI | 178 | 178 | 178 |
| 0.001 | NI | 135 | 135 | 135 |

(a)

| $d_{in}$ | Max | 60 | 50 |
|---|---|---|---|
| NI | 175 | 175 | 196 |

(b)

Table 3: (a) Results for *g*=2.0, *h*=0.02 and *T*=4, image pictured on Fig. 2.a. (Max. means the whole image); (b) Inner window reduction and performance for the image 2.b, *g*=2.0, *h*=0.02 and *T*=4 (see Fig. 2.d.).

## 6.3 Segmentation Approach

In this section, we demonstrate the utility of using GVF plus automatic snake initialization approaches. In this example, we take Fig. 2.c and isolate the darker object, which we assume that it is not an object of interest (Fig. 4.a). In a more complex image, the bounding box pictured might contain artifacts, noise, etc.

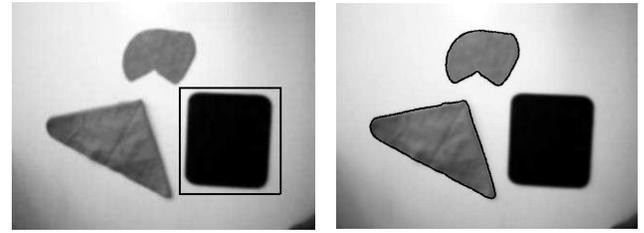

(a)          (b)

Figure 4: (a) Original image with bounding box.
(b) Final result with extracted boundaries of interest.

Then, we can consider that bounding box as a "hole" in the image and define a multiply connected domain holding the targets. Now, we can take advantage of the fact that the obtained curve is "close" to the desired boundary and use the GVF for multiply connected domains, with *g* = 0.2, *h* = 1.0 to get the new image force. This solution is not computational expensive (NI = 138, CPU time 7seg), and the obtained result is enough extended to allow snakes flow into the boundaries (Fig. 4.b). Besides, the diffusion-reaction effects are more local if the number of iterations is lower, which means that in a neighborhood of a high contrast region the initial field is more extended than mix with that one diffused from distant locations.

## 7. Conclusion and Future Works

In this paper we analyze the GVF models with respect to the parameter's selection. The number of iterations gets larger as *T* increases. The computational cost can be reduced by window methods, but we have to be careful with precision of the obtained result. Besides, advantages of GVF plus automatic initialization methods were highlighted. In this case, the number of iterations can be reduced in order to improve the field only closer the desired boundary. As further investigations, we should analyze the utility of the theory of functions of *bounded variation*.